\documentclass{article}
\usepackage{ijcai22}

\usepackage{times}
\usepackage{soul}
\usepackage{url}
\usepackage[hidelinks]{hyperref}
\usepackage[utf8]{inputenc}
\usepackage[small]{caption}
\usepackage{graphicx}
\usepackage{amsmath}
\usepackage{amsthm}
\usepackage{booktabs}
\usepackage{algorithm}
\usepackage{algorithmic}
\usepackage{amssymb}
\usepackage{subcaption}
\def\clip{\operatorname{clip}}

\title{Adventurer: Exploration with BiGAN for Deep Reinforcement Learning}

\author{
Yongshuai Liu
\And Xin Liu
\affiliations
University of California, Davis
\emails
\{yshliu, xinliu\}@ucdavis.edu,
}

\begin{document}

\maketitle

\begin{abstract}
  Recent developments in deep reinforcement learning have been very successful in learning complex, previously intractable problems. Sample efficiency and local optimality, however, remain signiﬁcant challenges. To address these challenges, novelty-driven exploration strategies have emerged and shown promising potential. Unfortunately, no single algorithm outperforms all others in all tasks and most of them struggle with tasks with high-dimensional and complex observations. In this work, we propose Adventurer, a novelty-driven exploration algorithm that is based on Bidirectional Generative Adversarial Networks (BiGAN), where BiGAN is trained to estimate state novelty. Intuitively, a generator that has been trained on the distribution of visited states should only be able to generate a state coming from the distribution of visited states.  As a result, novel states using the generator to reconstruct input states from certain latent representations would lead to larger reconstruction errors. We show that BiGAN performs well in estimating state novelty for complex observations. This novelty estimation method can be combined with intrinsic-reward-based exploration. Our empirical results show that Adventurer produces competitive results on a range of popular benchmark tasks, including continuous robotic manipulation tasks (e.g. Mujoco robotics) and high-dimensional image-based tasks (e.g. Atari games).
\end{abstract}

\section{Introduction}
Reinforcement learning (RL) has achieved impressive success on a variety of tasks~\cite{liu2021clara,liu2021cts2}, such as controlling simulated robots~\cite{schulman2015trust} and operating the AlphaGo system~\cite{silver2016mastering}. These successes, however, are mostly in the realm of easy-to-learn tasks~\cite{liu2021resource,liu2020constrained}: significant barriers remain to applying RL in hard-to-learn applications, such as tasks with sparse rewards and high-dimensional state space~\cite{liu2024towards,liu2023constrained}, which are the key challenges for current RL techniques to achieve high sample efficiency and globally optimal.

Research has shown that efficient exploration can significantly improve sample efficiency and escape locally optimal solutions in order to find a globally optimal solution~\cite{halev2024microgrid}. Exploration in RL encourages the agent to visit states that have not been visited (enough) in order to gather better trajectory data. 
Various strategies have been developed to encourage efficient exploration~\cite{liu2022farsighter}. Classical exploration methods, such as $\epsilon$-greedy DQN~\cite{mnih2015human} and Gaussian noise~\cite{lillicrap2015continuous}, add random noise to the output actions so that the probability that an agent visits any given action from each state is non-zero. Random exploration will eventually learn the optimal policy by doing blind searches but suffers from low sample efficiency. Since the agent does not remember where it has previously explored, it may repeatedly try actions and states it has already visited.

Recently, novelty-driven exploration algorithms with intrinsic rewards \cite{bellemare2016unifying,ostrovski2017count,zhao2019curiositydriven,tang2017exploration,fu2017ex2,burda2018exploration,badia2020never,dai2022diversity,bigazzi2022focus} have been developed; they perform much more efficiently than classical random exploration. By treating intrinsic rewards as an exploration bonus, the agent is encouraged to visit novel states -- intrinsic rewards are likely to be higher in novel states than frequently visited ones. Different methods have been developed to measure the degree of `novelty'. Count-based bonuses have been shown to provide substantial progress in both tabular-based and non-tabular-based RL methods~\cite{bellemare2016unifying,ostrovski2017count,zhao2019curiositydriven,tang2017exploration,fu2017ex2}. 
Other methods estimate forward prediction models and use the prediction error as the intrinsic reward~\cite{burda2018exploration}. Yet, no \textbf{single algorithm outperforms all others}~\cite{Taiga2020On} and most of them perform well \textbf{only on smaller tasks}, as stated in~\cite{fu2017ex2,badia2020agent57}. Therefore, it is still an \textbf{open question} how to best evaluate ``novelty'' under diverse environments, especially with \textbf{high-dimensional} and \textbf{complex} observations.

In this paper, we propose Adventurer, a novelty-driven exploration strategy based on Bidirectional Generative Adversarial Networks (BiGAN). The BiGAN structure allows us to efficiently estimate the novelty of complex high-dimensional states, e.g., image-based states. Intuitively, a BiGAN that has been well-trained on the visited states should only be able to reconstruct -- generate -- a state from the distribution of visited states. Specifically, 
for a given input state, we obtain its latent representation from the encoder and then reconstruct it from the BiGAN generator. The reconstruction error should be small for
the frequently visited states and states close to them, and large for the less-visited or new (novel) states.
In practice, we estimate state novelty by combing `pixel' level reconstruction error and `feature' level discriminator feature matching error which complement each other to provide a more accurate novelty estimation. 
This BiGAN-based novelty estimation component can work with any policy optimization algorithm as an intrinsic-reward-based exploration bonus. In summary, we make the following contributions: 
\begin{itemize}
    \item We develop a BiGAN-based scheme to estimate state novelty by combing `pixel' level reconstruction error and `feature' level discriminator feature matching error. 
    \item Our estimate method has great potential for tasks with \textbf{high-dimensional observations}. Moreover, this novelty estimation algorithm can be easily integrated into any policy optimization algorithm.
    \item  Empirical results show that the proposed method is scalable and achieves competitive performance on both continuous robotic manipulation tasks (Mujoco robotics) and high-dimensional image-based tasks (Atari games).
\end{itemize}

\section{Related Work}
Exploitation versus exploration is a critical topic in reinforcement learning. Efficient exploration can improve sample efficiency and lead to better policies. 
While much progress has been made in how to exploit most efficiently to achieve the best long-term returns, exploration remains an open problem~\cite{hao2023exploration} in modern RL algorithms.
The most common approach to increasing exploration of an environment is to augment the environmental (\textbf{extrinsic}) reward with an exploration bonus (\textbf{intrinsic} reward) that encourages the agent to explore more.
The policy is then trained with a reward composed of both extrinsic and intrinsic rewards. To estimate the intrinsic rewards, there are two main approaches: count-based and prediction-based.

\textbf{Count-based explorations}~\cite{bellemare2016unifying,van2016conditional,zhao2019curiositydriven,tang2017exploration,fu2017ex2} estimate a pseudo-count of a state has been visited. This is used to give states with little visited time a high exploration bonus; typically, the intrinsic reward is defined as $1/n(s)$ or $1/\sqrt{n(s)}$, where $n(s)$ is the number of times a state $s$ has been visited.

Some research \cite{bellemare2016unifying,ostrovski2017count,zhao2019curiositydriven} uses density models to approximate the visit frequency of states and then derive a pseudo-count from this density model. 
For example, in~\cite{bellemare2016unifying}, pseudo-counts are estimated with a Context Tree Switching (CTS)~\cite{bellemare2014skip} model.
To improve the scalability of CTS,~\cite{ostrovski2017count} improved the approach by training a model of PixelCNN~\cite{van2016conditional}. Moreover, the density model can also be a Gaussian Mixture Model~\cite{zhao2019curiositydriven}. 
$EX^{2}$~\cite{fu2017ex2} estimates implicit density by considering how easily a given state is distinguished from other visited states by a discriminatively trained classiﬁer. $\#$ exploration~\cite{tang2017exploration} considers the problem of high-dimensional states by mapping states into shortened hash codes so that the occurrences of states become trackable by using a hash collision. However, the mapping fails with more complex observations when dissimilar observations may be mapped to identical hashes~\cite{weng2020exploration}.

\textbf{Prediction-based approaches} use prediction error to estimate state novelty and thus the intrinsic reward. 
Random Network Distillation (RND)~\cite{burda2018exploration} receives intrinsic exploration bonuses from a neural network predicting error with a fixed randomly initialized neural network. RND fits the prediction neural network to the fixed randomly initialized neural network if an input state has been visited. The motivation is that, given a state, the prediction error should be lower if similar states have been visited many times in the past. Empirical results show that RND performs better in tasks with a non-episodic setting where novelty can be learned across multiple episodes.

Our method can be considered as a prediction-based method. In comparison, our method performs better in general in more complex scenarios where traditional count-based and prediction-based methods suffer in.

\noindent\textbf{Open Question:} While exploration  has been much studied in the literature,  {no single algorithm outperforms all others} in all environments~\cite{hao2023exploration,Taiga2020On}. Therefore, there is a strong need for better and more diverse exploration techniques. Specifically, 
most existing work performs well on smaller tasks but suffers in tasks with \textbf{complex observations}~\cite{bellemare2016unifying,badia2020never}.
To address this challenge, the BiGAN structure in our work shows great benefit for estimating novelty, especially for complex high-dimensional states.

\section{Preliminaries}
\textbf{Markov Decision Process:}
A Markov Decision Process (MDP) is defined by the tuple $\left(\mathit{S},\mathit{A}, \mathit{R},\mathit{P},\mathit{\mu}, \mathit{\gamma} \right )$~\cite{liu2021policy,liu2020ipo}, where $\mathit{S}$ is the set of states, $\mathit{A}$ is the set of actions, $\mathit{R}$ is the reward function under a state and action pair, $\mathit{P}$ is the transition probability function from state $s$ to state $s^{'}$ with taking action $a$, $\mathit{\mu}$ is the initial state distribution and $\mathit{\gamma}$ is the reward discount factor. A policy $\pi(a \mid s)$ is the probability of taking action $a$ in state $s$. Policy $\pi$ is usually written as $\pi_{\theta}$ to emphasize its dependence on the parameter $\theta$. The goal of an MDP is to learn a policy $\pi_{\theta}$ which maximizes the discounted cumulative reward. This objective is denoted as
$$
    \max_{\theta}~J_{R}^{\pi_{\theta}} = \mathbb{E}_{\tau \sim \pi_{\theta}}[\sum_{t=0}^{\infty}\mathit{\gamma}^{t}\mathit{R}(s_{t},a_{t},s_{t+1})],
$$
where $s_t$ is the state in timestep $t$, $a_t$ is the action in timestep $t$, $\tau = (s_{0}, a_{0},s_{1}, a_{1}... )$ denotes a trajectory, and $\tau \sim \pi_{\theta}$ means that the distribution over trajectories is following policy $\pi_{\theta}$.
The value function of state $s$ is
$$
    V^{\pi_{\theta}}{(s)} = \mathbb{E}_{\tau \sim \pi_{\theta}}[\sum_{t=0}^{\infty}\gamma^{t}\mathit{R}(s_{t},a_{t},s_{t+1})\mid s_{0}=s].
$$
The  action-value function of state $s$ and action $a$ is 
$$
    Q^{\pi_{\theta}}{(s,a)} = \mathbb{E}_{\tau \sim \pi_{\theta}}[\sum_{t=0}^{\infty}\gamma^{t}\mathit{R}(s_{t},a_{t},s_{t+1})\mid s_{0}=s, a_{0}=a].
$$
and the advantage function is 
\begin{equation}\label{eq:advantage}
    A^{\pi_{\theta}}{(s,a)} = Q^{\pi_{\theta}}{(s,a)} - V^{\pi_{\theta}}{(s)}
\end{equation}

\textbf{Proximal Policy Optimization:}
Many RL methods exist to solve an MDP; our exploration strategy can be integrated with any of them. In practice, we take advantage of Proximal Policy Optimization (PPO)~\cite{schulman2017proximal}. PPO is an example of a policy gradient method; it approximates the objective by a clipped surrogate objective. PPO is able to achieve monotonic improvement when updating policy with first-order optimization (e.g. Adam). The PPO objective is

\begin{equation}
\label{eq:PPO}
\resizebox{\hsize}{!}{
  $\max_{\theta}~L^{CLIP}(\theta ) = \mathbb{E}_{t}[\min(r_{t}(\theta)A_{t}, \clip(r_{t}(\theta),1-\epsilon ,1+\epsilon)A_{t})],$
  }
\end{equation}
where $r_t(\theta) = \frac{\pi_{\theta}(a_{t} \mid s_{t})}{\pi_{\theta_{old}}(a_{t} \mid s_{t})}$, $A_{t}$ is the advantage function (Eq.~\ref{eq:advantage}), $\clip(\cdot)$ is the clip function and $r_{t}(\theta)$ is clipped between $\left [ 1-\epsilon, 1+\epsilon  \right ]$. 

\section{Adventurer: Exploration with BiGAN}
We first describe the model used to estimate the state novelty during training. We then describe how to use this model as an intrinsic reward in PPO.

\subsection{State Novelty Estimation}\label{sec:estimation}
\begin{figure*}[t]
    \centering
    \includegraphics[width=0.85\textwidth]{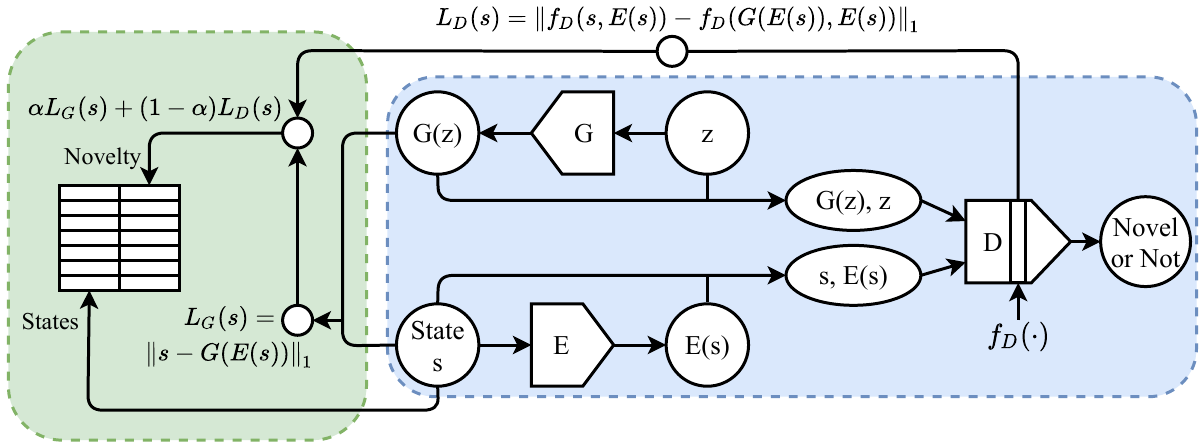}
    \caption{Adventurer architecture. Blue: the structure of BiGAN; Green: the workflow of state novelty estimation. The BiGAN is trained on visited states. The novelty of an input state $s$ is estimated by $\alpha L_{G}(s)+(1-\alpha)L_{D}(s)$.}
    \label{fig:adventurer}
\end{figure*}

We propose to use Bidirectional Generative Adversarial Networks (BiGAN)~\cite{donahue2017adversarial} to estimate state novelty. As shown in Fig.~\ref{fig:adventurer} (the blue part), BiGAN extends the GAN framework by adding an encoder $E(s)$ which learns a mapping from state to a latent representation. The encoder learns the inverse of the generator $E = G^{-1}$. The generator maps a source of random vector $z \sim  P_{Z}$ to a synthetic state $\widehat{s} = G(z)$. The BiGAN learns a mapping from the latent space to the visited state space and vice versa, simultaneously.  The discriminator learns to classify a $(s, E(s))$ or $(G(z), z)$ pair, instead of only learning to classify input state samples. It produces an estimation of whether the state $s$ is sampled from the visited state distribution $p_{S}$ or generated by $G(z)$.

The purpose of BiGAN training is to learn a discriminator to reliably distinguish whether a state $s$ is novel or not and to use this discriminator to train a good generator that fits the distribution of visited state by trying to fool the discriminator. In other words, the discriminator $D(s, E(s))$ is trained to maximize the probability of assigning the
not novel label to the visited states and assigning the novel label to synthetic states from generator $G(z)$. 
Discriminator, generator, and encoder play the two-player minimax game~\cite{donahue2017adversarial,schlegl2017unsupervised}:
\begin{equation}\label{eq:bigan}
   \begin{aligned}
  &\min_{G,E} \max_{D}V(D,G,E)= \\
  &\mathbb{E}_{s \sim p_{S}}[\mathbb{E}_{E(s) \sim p_{E(\cdot \mid s)}}[log D(s,E(s))]]+\\
  &\mathbb{E}_{z \sim p_{Z}}[\mathbb{E}_{G(z) \sim p_{G(\cdot \mid z)}}[1-log D(G(z),z)]]
\end{aligned} 
\end{equation}
where $p_{S}$ is the distribution over all visited states, $p_{Z}$ is the distribution over the latent representation, and $p_{E(\cdot \mid s)}$ and $p_{G(\cdot \mid z)}$ are the distributions induced by the encoder and generator respectively.

BiGAN estimates state novelty as follows:
we use the generator to reconstruct an input state from a certain latent representation learning from the encoder. The input states that have larger reconstruction errors are more likely to be novel. A BiGAN, which is trained only on visited states, forces the generator to learn the manifold of visited states distribution. The generator should be only able to generate such a state which is similar to the visited states. Moreover, since the encoder learns the inverse of generator $E=G^{-1}$,
ideally, when a frequently visited state $s$ is encoded by the encoder $E(s)$ and then reconstructed by the generator $G(E(s))$, its reconstruction error between the input state and the reconstructed state should be zero, since $ \left \| s - G(E(s)) \right \| = \left \|s - G(G^{-1}(s))  \right \| = \left \| s- s \right \| = 0$; On the contrary, when a novel state ${s}'$ is reconstructed by the encoder $E({s}')$ and then generator $G(E({s}'))$, the reconstruction error $ \left \|{s}' - G(E({s}')) \right \|$ will be higher.

We then deﬁne a novelty score function $B(s)$, inspired by anomaly detection in~\cite{schlegl2017unsupervised}, to quantify how novel a state $s$ is.
The score function is a combination of two components: reconstruction error $L_{G}(s)$ and discriminator-based error $L_{D}(s)$: 
\begin{equation}\label{eq:novel}
    B(s)=\alpha L_{G}(s)+(1-\alpha) L_{D}(s) 
\end{equation}
where $L_{G}(s) = \left \| s-G(E(s)) \right \|_{1}$ enforces the similarity between input state $s$
and the reconstructed state $G(E(s))$. $L_{D}(s) = \left \| f_{D}(s,E(s)) - f_{D}(G(E(s)),E(s)) \right \|_{1}$ takes into account the discriminator, where $f_{D}(\cdot)$ is the output of an intermediate layer of the discriminator as in Fig.~\ref{fig:adventurer}. This considers whether the reconstructed state has similar features as the visited states in the discriminator. 

State novelty estimation requires the novelty score to be small not only for states that have been visited explicitly but for states that are near the visited states. Moreover, the novelty score between (near) visited states and non-visited states should be distinguishable. Neither $L_G$ nor $L_D$ alone can achieve this. $L_G$ measures the novelty on the `pixel' level. As shown in Sec.~\ref{sec:mr}, if we only consider the $L_G$, for near-visited states, the generator cannot reconstruct well which leads to large $L_G$.
On the other hand, $L_D$ measures the novelty on the `feature' level. It tells whether a state fits the learned distribution of visited states. Even though a near-visited state has not been explicitly visited, the $L_D$ remains small since it has similar features to visited states. However, if we only consider the $L_D$, the novelty scores are close for visited, near-visited, and non-visited states, which cannot show an evident difference.
Adventurer considers both $L_G$ on the `pixel' level and $L_D$ on the `feature' level. They complement each other to provide a more accurate novelty estimation. 

Using BiGAN, we learn a generative model of the visited states. It also enables us to model complex high-dimensional states, e.g. image-based states, efficiently. The frequently visited states and states close to them have smaller $B(s)$. States that are less frequently visited or never seen before have larger values of $B(s)$. We validate this idea in the experiment Sec.~\ref{sec:val}.

\subsection{Combining Intrinsic Reward with Extrinsic Reward}
\label{combine}
We treat $B(s)$ as the exploration bonus function. One issue with using $B(s)$ as an intrinsic reward is that the scale of the reward can differ greatly from the extrinsic reward and it may vary at different time points. In order to keep the intrinsic rewards $r^{i}(s)$ on a same scale of extrinsic rewards $r^{e}$ and reduce the variation, we normalize the exploration bonus $B(s)$ as 
\begin{equation}\label{eq:normalization}
    r^{i}(s) = \frac{B(s) - \mu(B(s)) + \mu(r^{e})}{\sigma (B(s))}
\end{equation}
where $\mu(r^{e})$, $\mu(B(s))$, $\sigma (B(s))$ is the running estimation of  extrinsic reward average, intrinsic reward average, and standard deviation, respectively.

Instead of directly augmenting the extrinsic rewards $r^{e}$ in each step with the intrinsic reward $r^{i}$ using $r_{t}= r^{e}_{t}+ \beta r^{i}_{t}$ at time $t$, we fit two advantage functions $A^{e}_{t}$, $A^{i}_{t}$ (Eq.~\ref{eq:advantage}) for the rewards, respectively. We then combine them as the augmented advantage function
$A_{t}= A^{e}_{t}+ \beta A^{i}_{t}$, where we use $A_{t}$ to optimize policy in PPO (Eq.~\ref{eq:PPO}). $\beta$ is a hyperparameter adjusting the balance between exploitation and exploration.
Learning the extrinsic and intrinsic rewards separately provides the flexibility of adjusting the exploration bonus~\cite{burda2018exploration,badia2020agent57}, as discussed in Appendix~\ref{sec:separate}. 
The novelty-driven deep RL agents are trained on the augmented advantage function $A_{t}$, while the policy performance is evaluated only on the long-term return of extrinsic reward $r^{e}_{t}$.

\subsection{The Adventurer Algorithm}
We now demonstrate how to incorporate the intrinsic reward into a full PPO agent to improve exploration efficiency. Pseudocode for Adventurer is shown in Algorithm~\ref{alg:adventurer}. Adding to PPO,
in each step, we calculate the state novelty score (line 7), which is the intrinsic reward that drives the exploration. After we get all samples in each epoch, we normalize the novelty scores (line 12) and update policy with the objective function Eq.~\ref{eq:PPO} (line 16) using the augmented advantage $A_t$ (line 15) instead of the extrinsic reward advantage. Moreover, we also update the BiGAN to fit the visited states distribution (line 17).

\section{Experiment}
\label{sec:exp}

In this section, we investigate the following questions:
\begin{itemize}
    \item Can BiGAN estimate state novelty with the novelty score $B(s)$ (Eq.~\ref{eq:novel}) and why is better?
    \item Does incorporate BiGAN-based intrinsic reward bring exploration benefit to typical policy optimization algorithms compared to baselines?
\end{itemize}

\subsection{BiGAN Validation}\label{sec:val}
\subsubsection{CIFAR-10}\label{sec:cifar10}
\begin{figure}[t]
    \centering
    \includegraphics[width=0.8\columnwidth]{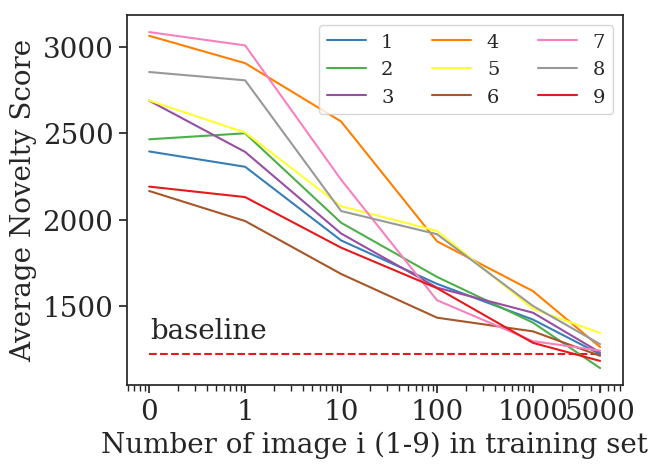}
    \caption{Novelty score on CIFAR-10. 
    The dashed line represents the novelty score that the BiGAN is trained only on image 0 and tested with image 0.
    }
    \label{fig:cifar}
\end{figure}

To validate if the BiGAN can be used to estimate state novelty with the novelty score $B(s)$, especially for tasks with complex high-dimensional states. We did an experiment on CIFAR-10~\cite{krizhevsky2009learning} inspired by~\cite{burda2018exploration}. We train a BiGAN on a dataset containing all training images with label 0 and partial images with a specified label $i$, where $i$ could be 1 to 9. The total number for image 0 is 5000, while the number of specified images $i$ varies from 0 to 5000. We then test the BiGAN with all test images of label $i$ and calculate the novelty score for each.
Intuitively, the training images 0 are the states we have visited most time. Images $i$ are the states we have not visited frequently. The results in Figure~\ref{fig:cifar} show that as the number of images $i$ increases in the training set, the average novelty score decreases. In other words, as the visited time increase for a given state $s$, the novelty score $B(s)$ decreases, which suggests that novelty score $B(s)$ can be used to estimate the state novelty.

\begin{algorithm}[th]
	\caption{Adventurer: novelty-driven exploration} \label{alg:adventurer}
	\begin{flushleft}
	Initialize policy network $\pi$ with parameters $\theta_{\pi}$;
	BiGAN network with parameters $\theta_{BiGAN}$;
    Novelty score reconstruction error scale $\alpha$;
	Exploration weight $\beta$;
	Total epoch number $L$;
	Total episodic number in each epoch $N$;
	Maximum step size for each episode $H$.
	\end{flushleft}
	
	\begin{algorithmic}[1]
	    \FOR {l = 0,...,L}
	    \FOR {n = 0,1,...,N}
	    \STATE Sample a random initial state $s_{0}$ from the environment.
	    \FOR {h = 0,1,...,H}
	    \STATE Sample action $a_{t}\sim\pi(a_{t} \mid s_{t})$.
	    \STATE Sample $s_{t+1}, r^{e}_{t} \sim p(s_{t+1}, r^{e}_{t} \mid s_{t},a_{t})$.
	    \STATE Calculate novelty score $B(s_{t+1})$ with Eq.~\ref{eq:novel}.
	    \STATE Store $(s_{t},a_{t},r^{e}_{t},B(s_{t+1}),s_{t+1})$ into optimization buffer.
	    \ENDFOR
	    \ENDFOR
	     \STATE Update intrinsic reward normalization parameter $\mu(r^{e})$, $\mu(B(s))$, $\sigma(B(s))$.
	    \STATE Normalize $B(s_{t+1})$ into intrinsic reward $r^{i}_{t}$ with Eq.~\ref{eq:normalization}.
	    \STATE Calculate extrinsic reward advantage $A^{e}_{t}$ with Eq.~\ref{eq:advantage}.
	    \STATE Calculate intrinsic reward advantage $A^{i}_{t}$ with Eq.~\ref{eq:advantage}.
	    \STATE Calculate augmented advantage $A_{t} = A^{e}_{t} + \beta A^{i}_{t}$.
	    \STATE Update the policy network parameters $\theta_{\pi}$ with Eq.~\ref{eq:PPO}.
	    \STATE Update the BiGAN parameters $\theta_{BiGAN}$ on visited states with Eq.~\ref{eq:bigan}.
		\ENDFOR
	\end{algorithmic}
\end{algorithm}

\subsubsection{Montezuma’s Revenge}\label{sec:mr}
This experiment is designed to evaluate novelty estimation in an RL setting.
Specifically, we show the BiGAN-based novelty estimation $B(s)$ is better than RND~\cite{burda2018exploration} and VAE~\cite{asperti2021survey} reconstruction error-based novelty estimation. It is also better than just using an individual component $L_{G}(s)$ or $L_{D}(s)$. 

\begin{figure*}[t]
    \centering
    \hfill
     \begin{subfigure}[t]{0.19\textwidth}
         \centering
         \includegraphics[width=\textwidth]{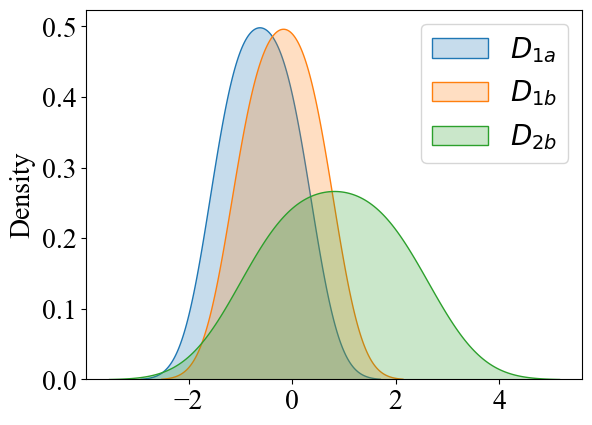}
         \caption{$B(s)$}
     \end{subfigure}
     \hfill
     \begin{subfigure}[t]{0.19\textwidth}
         \centering
         \includegraphics[width=\textwidth]{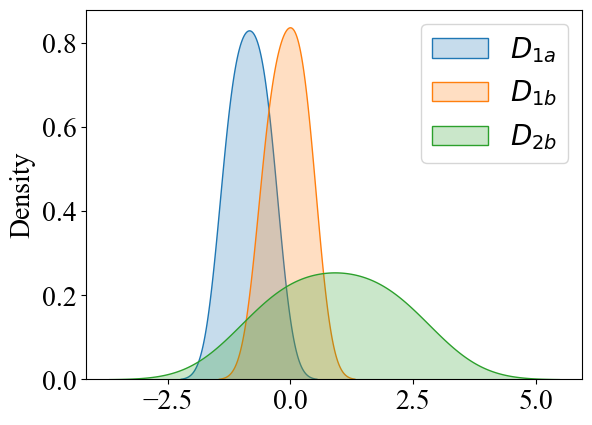}
         \caption{$L_G(s)$}
     \end{subfigure}
    \hfill
    \begin{subfigure}[t]{0.19\textwidth}
         \centering
         \includegraphics[width=\textwidth]{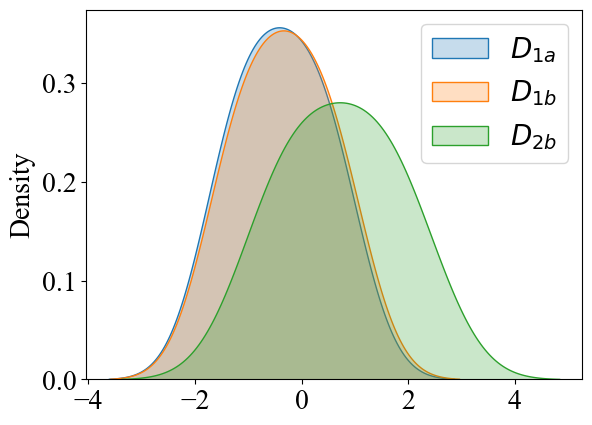}
         \caption{$L_D(s)$}
     \end{subfigure}
     \hfill
     \begin{subfigure}[t]{0.19\textwidth}
         \centering
         \includegraphics[width=\textwidth]{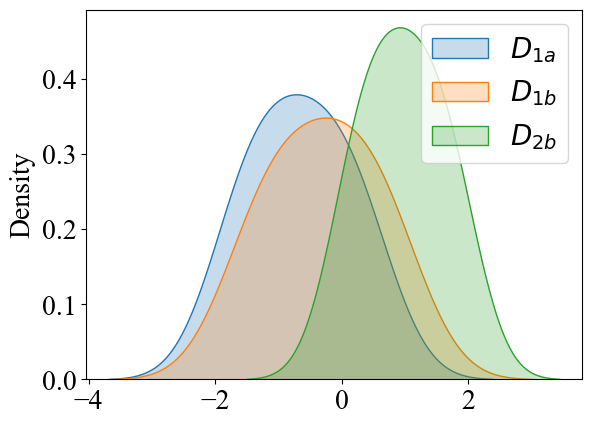}
         \caption{RND}
     \end{subfigure}
     \hfill
     \begin{subfigure}[t]{0.19\textwidth}
         \centering
         \includegraphics[width=\textwidth]{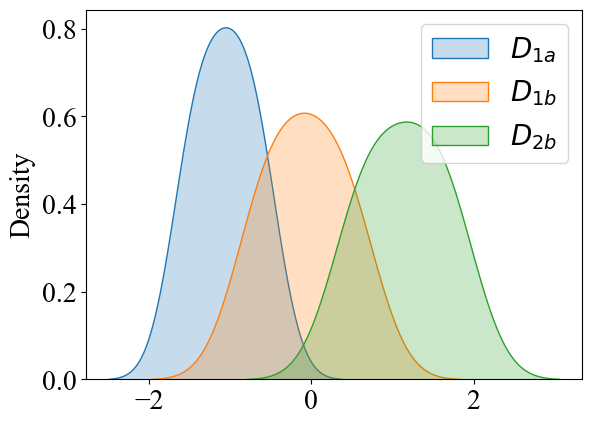}
         \caption{VAE}
     \end{subfigure}
     \hfill
    \caption{Setting 1: the novelty score distributions for novelty estimators $B(S)$, $L_G(s)$, $L_D(S)$, RND,and VAE, which are trained with observations $D_{1a}$.}
    \label{fig:point3}
\end{figure*}

We randomly sample two sets of different observations $D_1$ and $D_2$ from a complex image-based Atari game, Montezuma’s Revenge. $D_1$ and $D_2$ are observations from two different rooms in Montezuma’s Revenge game. Therefore, observations in each set are similar to each other but differ in different sets. We further divide $D_1$ to $D_{1a}$ and $D_{1b}$ and divide $D_2$ to $D_{2a}$ and $D_{2b}$. The number of observations in  $D_{1a}$ and $D_{2a}$ are equal. We consider two settings to evaluate novelty estimation and show the results in Figure~\ref{fig:point3}.
\begin{itemize}
    \item Setting 1: we estimate the novelty scores of observations $D_{1a}$, $D_{1b}$, and $D_{2b}$ on estimators trained with $D_{1a}$ and quantify the estimators with $D_{KL}(D_{1b} || D_{1a}) - D_{KL}(D_{2b} || D_{1a})$, where $D_{KL}$ is the Kullback–Leibler (KL) divergence. The observations $D_{1a}$ and $D_{1b}$ are from the same set $D_{1}$. They are treated as visited observations. $D_{2b}$ is from $D_{2}$ which is total new to the novelty estimators. Thus, we want the distribution between visited observations as the same as possible. On the contrary, the distribution of non-visited observations is as different as possible. The combined KL divergence is smaller the better. In all novelty estimation methods, the distributions of normalized novelty scores show that most novelty scores of $D_{1a}$ and $D_{1b}$ are lower than novelty scores of $D_{2b}$. This demonstrates that all novelty estimation methods are reasonable. BiGAN-based $B(s)(\alpha=0.9)$ achieves the best performance (Appendix~\ref{sec:hyper} will discuss hyperparameters selection). The distributions of $D_{1a}$ and $D_{1b}$ overlap and the distribution of $D_{2b}$ is very different. RND is second only to $B(s)(\alpha=0.9)$. VAE shows much difference between the score distribution of  $D_{1a}$ and $D_{2b}$ observations. However, the distance between the distribution of  $D_{1a}$ and $D_{1b}$  observations is the highest as well. $L_{G}(s)$ obtains large difference between distribution of  $D_{1a}$ and $D_{2b}$  but the distance between distribution of  $D_{1a}$ and $D_{1b}$  is a little bit large. On the contrary, $L_{D}(s)$ obtains small difference between  $D_{1a}$ and $D_{1b}$  , but the distance over distribution of  $D_{1a}$ and $D_{2b}$  is small.
 \item Setting 2: we estimate the novelty scores of observations $D_{1a}$, $D_{1b}$, $D_{2a}$, and $D_{2b}$ on estimators trained with $D_{1a}$ and $D_{2a}$ and quantify the estimators with $D_{KL}(D_{1b} || D_{1a}) + D_{KL}(D_{2b} || D_{2a})$. Since the novelty estimators are learned on observations from both  $D_{1a}$ and $D_{2a}$ and the number of observations is equal. So all the test observations are treated as visited observations for the novelty estimators. Thus, we want the combined KL divergence to be small as well. For each estimation method, the distributions of normalized novelty scores are close. Moreover, the distributions of BiGAN-based score $B(s)(\alpha=0.9)$ are closest among all methods, they overlap with each other. $L_{D}(s)$ performs similar with $B(s)(\alpha=0.9)$  and then RND and $L_{G}(s)$. VAE performs worse, and the differences between each distribution are relatively large.
\end{itemize}

In both settings, the experiment demonstrates that BiGAN-based novelty score $B(s)(\alpha=0.9)$ gives the best novelty estimation compared to baselines of RND, VAE-based methods, and each individual component $L_{G}(s)$ or $L_{D}(s)$. The score is low for visited observations and high for novel observations. Moreover, the difference is the most evident. 

As we observed, neither reconstruction error (e.g. $L_G$, VAE) nor $L_D$ alone is sufficient to estimate state novelty. 
As discussed in Sec~\ref{sec:estimation}, Adventure 
requires the novelty score to be small not only for states that have been visited explicitly but for states that are near the visited states.
If we only consider the reconstruction error (e.g., $L_G$, VAE), the novelty score will be relatively large for both (near) visited states and non-visited states, as shown in Fig.~\ref{fig:point3}. The reason is that for states that have not been visited but are near the visited states, the $L_G$ and VAE cannot reconstruct well which leads to large reconstruction errors. If we only consider the $L_D$, the novelty scores
are close for visited, near-visited, and non-visited states. Adventurer considers a feature matching error $L_D$ additionally with $L_G$. $L_D$ takes into account if a state has similar features to the visited ones and reduces the negative effect of only considering reconstruction error. 
$L_{G}(s)$ performs better in detecting novel observations and $L_{D}(s)$ outperforms in reducing the (near) visited observations novelty score. Therefore, it is desirable to combine $L_{G}(s)$ and $L_{D}(s)$ together to  $B(s)$.

Moreover, BiGAN learns a better generative model for novelty estimation in complex high-dimensional scenarios, as also suggested in~\cite{zenati2018adversarially}. 
GANs are empirically superior as deep generative models compared to AEs~\cite{bergmann2018improving}. AEs are easy to produce blurry reconstructions for images and dependencies between pixels are not properly modeled.
Thus the reconstruction errors are relatively large for both visited and novel observations. 
A similar result exists in the evaluation of VAE-based anomaly detection~\cite{zenati2018adversarially}. The performance is worse than the BiGAN-based method as well. 

Theoretically, traditional GAN can also be used to reconstruct test observations and estimate novelty. However, it requires some optimization steps for every new generator input, which results in poor test-time performance~\cite{pang2021deep}. The benefit of BiGAN compared to typical GAN is that BiGAN learns the embedding simultaneously with the discriminator and generator. It can easily reconstruct new test observations by using $G(E(s))$. We can also use the discriminator score of GAN to estimate state novelty.  In fact, GAEX~\cite{hong2019generative}, which is a GAN-based method, works in that way. GAEX works similarly with $L_D$. The novelty score for (near) visited states and non-visited states are close and do not perform as well as ours, as shown in the evaluation Sec.~\ref{sec:performance}.

\subsection{Exploration Performance}
\subsubsection{Tasks}
Our exploration strategy can be used on both continuous actions and large complex state spaces.
The robotic manipulation tasks are in low-dimension but the action space is continuous.  On the other hand, the Atari games have discrete actions, but the states are high-dimensional images, which makes the exploration harder. We perform the experiments on both robot manipulation tasks and Atari games. In all environments, we consider sparse rewards which is the key exploration challenge in RL.

Specifically, we consider two challenging robot control tasks: FetchPickAndPlace and HandManipulateBlock. Moreover, we also conduct our experiments on hard-to-explore Atari games: Montezuma’s Revenge, Gravitar, and Solaris.

\subsubsection{Setting}
Intrinsic-reward-based exploration faces a fundamental limitation of vanishing intrinsic rewards~\cite{ecoffet2021first,badia2020never}: as an agent explores the environment and becomes familiar with some local areas after a number of steps, the agent loses the exploration bonus and is unable to return to novel areas. As a result, the policy it learns is driven by extrinsic rewards only.

Although the intrinsic reward vanishing problem is not the main focus of the paper, we can address it and further improve the performance with episodic memory if the environment is resettable. We can use episodic memory to remember novel states that have previously been visited. 
In subsequent episodes, the agent ﬁrst returns to a novel state (without exploration) stored in the memory, then explores from it. 
To realize it, we require the environment to be resettable: given a state, the agent can return to it without exploration. 
Whether an environment is resettable depends on applications. For example, this resettable assumption is reasonable in most simulation-based environments, e.g. games.

Our Adventurer algorithm can work \textbf{with or without} the episodic memory depending on whether the environment is resettable.
By adding resettability, we can further improve the exploration performance. Even if the environment cannot support resettability, our pure BiGAN-based intrinsic-reward exploration can still achieve competitive results compared with the pure exploration SOTA methods.

For Adventurer and all baselines, we run the agent with and without a resettable premise. It means if the resettable premise holds, the agent can return to a novel initial state stored in the episodic memory. Otherwise, the agent chooses a random initial state given by the environment for each episode. A detailed discussion for using the resettable premise is provided in the appendix~\ref{sec:reset}.

\begin{figure*}[t]
    \centering
    \begin{subfigure}[t]{0.32\textwidth}
        \centering
         \includegraphics[width=\textwidth]{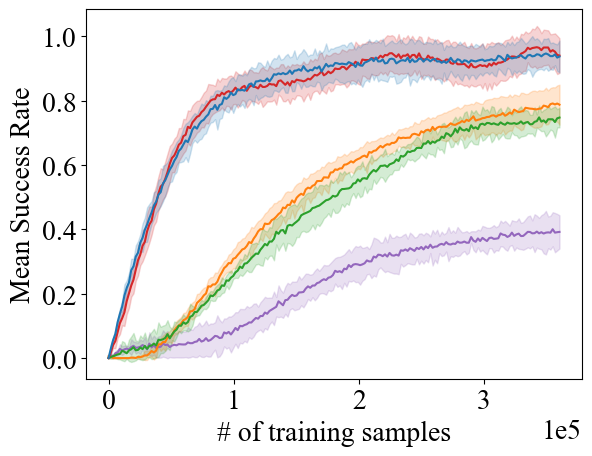}
         \caption{FetchPickAndPlace}
         \label{fig:fetch}
    \end{subfigure}
    \begin{subfigure}[t]{0.32\textwidth}
         \centering
         \includegraphics[width=\textwidth]{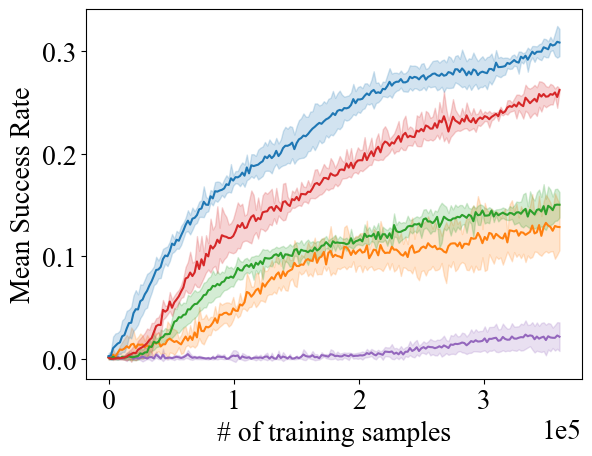}
         \caption{HandManipulateBlock}
         \label{fig:hand}
     \end{subfigure}
     \begin{minipage}[t]{0.9\textwidth}
         \centering
         \includegraphics[width=\textwidth]{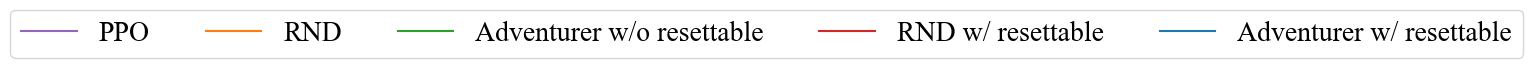}
    \end{minipage}
    \caption{The mean success rate for both Fetch arm and Shadow Dexterous hand tasks. The x-axis is the number of training samples.}
    \label{fig:robot}
\end{figure*}

\subsubsection{Baselines}
We compare Adventurer to PPO without an exploration bonus; RND~\cite{burda2018exploration}, which uses the forward prediction errors as intrinsic rewards; VAE-based method, which uses the reconstruction errors as intrinsic rewards and  GAEX~\cite{hong2019generative}, which uses the discriminator scores of GAN intrinsic rewards.

We note more recent papers on exploration, such as NGU~\cite{badia2020never}, Agent57~\cite{badia2020agent57} or Go-explore~\cite{ecoffet2021first},  integrate state novelty estimation with other methods, such as hyperparameter tuning or novel state resetting, to further improve the overall performance. These methods use state novelty estimation as a fundamental building block. Adventurer focuses on state novelty estimation so that our method can replace the existing state novelty estimation block in those methods.
Therefore, we only compare Adventurer with existing pure state novelty estimation methods in our evaluations. Moreover, as shown in~\cite{Taiga2020On,burda2018exploration}, RND achieves the SOTA performance compared with other pure novelty estimation algorithms in most tasks, although not all. Therefore, we choose it as the baseline. 

\subsubsection{Performance}\label{sec:performance}
We run each experiment 6 times with different random seeds and show the average performance. The shaded area is the standard deviation, as shown in Figure~\ref{fig:robot} and~\ref{fig:atari}. 
The discussion of hyperparameters (e.g. $\alpha$ and $\beta$, etc.) is provided in the appendix~\ref{sec:hyper}.

For robotic manipulation tasks, we compare the mean success rates, as shown in Figure~\ref{fig:robot}. The PPO performs worse in both cases since PPO uses random exploration which is inefficient.
In the simple FetchPickandPlack environment, Adventurer and RND achieve similar performance and they both outperform PPO. When the agent achieves around 40\% success rate, the PPO requires around 400,000 samples, while, Adventurer and RND without resettable only need around 100,000 samples.
Furthermore, the resettable premise can significantly improve performance.
Promising results are also achieved in the HandManipulateBlock environments. HandManipulateBlock is more complex than FetchPickandPlack. In the environment, Adventurer outperforms RND for both resettable and non-resettable conditions. Our experiments show that Adventurer achieves almost 100\% success rate in FetchPickandPlack and performs much better in HandManipulateBlock.

Figure~\ref{fig:atari} compares the game score of different algorithms for hard-to-explore Atari games: Montezuma’s Revenge, Gravitar, and Solaris.
In Montezuma’s Revenge and Gravitar, PPO performs worse. Adventurer outperforms RND by more than 20$\%$ in both resettable and non-resettable settings. RND is better than VAE and GAEX in the non-resettable setting. Moreover, the learning curve variance for VAE is much higher than others.

RND estimates the state novelty by distilling a ﬁxed random network (target network) into another predictor network. For each state, the target network produces random features of the state. The predictor is then trained to fit the features. However, random features in the target network may be insufﬁcient to represent the diverse environments. VAE uses the reconstruction error as the intrinsic reward which is large for all states and makes the policy hard to converge, thus the learning curve variance is large. The GAEX uses the discriminator score as the intrinsic reward. As we discussed in Sec.~\ref{sec:mr}, the scores
are close for visited, near-visited, and non-visited states which is similar to $L_D$. Thus performance improvement is limited.

\begin{figure*}[t]
    \centering
    \hfill
     \begin{subfigure}[t]{0.32\textwidth}
         \centering
         \includegraphics[width=\textwidth]{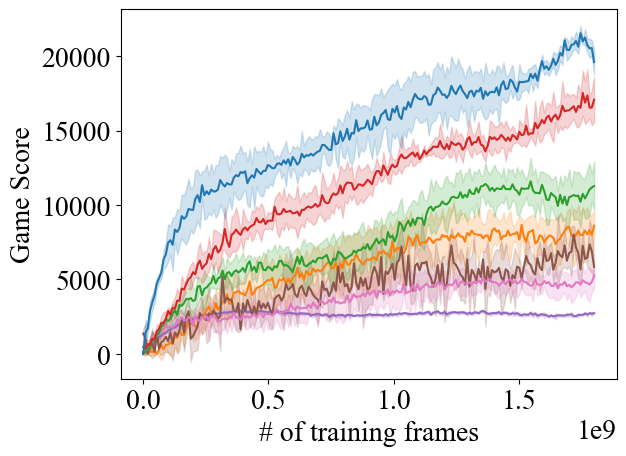}
         \caption{Montezuma’s Revenge}
     \end{subfigure}
     \hfill
     \begin{subfigure}[t]{0.32\textwidth}
         \centering
         \includegraphics[width=\textwidth]{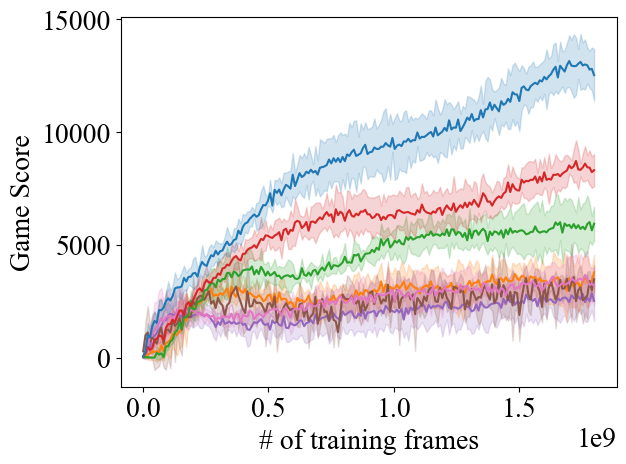}
         \caption{Gravitar}
     \end{subfigure}
    \hfill
    \begin{subfigure}[t]{0.32\textwidth}
         \centering
         \includegraphics[width=\textwidth]{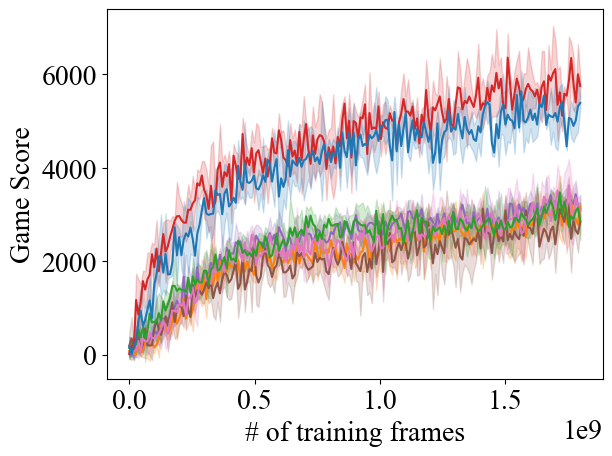}
         \caption{Solaris}
     \end{subfigure}
     \hfill
     \begin{minipage}[t]{0.9\textwidth}
         \centering
         \includegraphics[width=\textwidth]{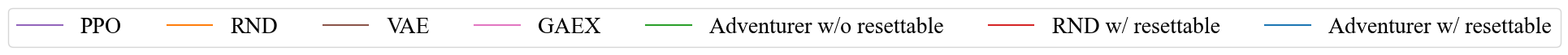}
    \end{minipage}
    \caption{The game score for Montezuma’s Revenge, Gravitar, and Solaris. The x-axis is the number of training frames.}
    \label{fig:atari}
\end{figure*}

In Solaris, Adventurer performs equivalently with RND in both settings and performs equivalently with other methods including PPO in a non-resettable setting. We conjecture the reason is that Solaris is a task with long-term delayed rewards, which is still a challenging problem for exploration in RL~\cite{hao2023exploration}.

Overall, Adventurer shows competitive performance on both continuous robotic manipulation tasks and high-dimensional state tasks in both resettable and non-resettable. 

\section{Discussion}
\label{sec:diss}
We present a novelty-driven exploration strategy for RL based on BiGAN, which shows great benefit in
complex high-dimensional observations, especially. The combination of `pixel' level reconstruction error and `feature' level discriminator feature matching error leads to more accurate novelty estimation at the cost of a fair amount of computation resources and time in the BiGAN training process.
How to best evaluate `novelty' in diverse environments is still an open question. There is no single algorithm that outperforms all others. 
Therefore, there is a strong need for better and more diverse exploration techniques, especially with high-dimensional and complex observations.

Adventurer can work \textbf{with or without} the resettable premise. Even if the environment cannot support resettability, our pure BiGAN-based exploration can still achieve competitive results compared with the pure exploration SOTA. When the environment is resettability, we can further improve exploration performance. 
As shown in Figure~\ref{fig:robot} and~\ref{fig:atari}, the resettability almost doubles the performance compared to pure intrinsic-reward-based exploration. 
Fortunately, in most simulation-based RL environments, the resettable assumption is typically supported. This also benefits policies that are learned in a simulator before transferring to their real-world applications. 
Even in the real world, the premise can be supported in some cases. For example, we can reset the states effortlessly in a go game. For cases that cannot be reset easily, we can learn a goal-conditioned policy~\cite{andrychowicz2017hindsight}, as suggested in~\cite{ecoffet2021first}, to help the agent return to the given novel states.

\section{Conclusion}
We propose Adventurer, a novelty-driven exploration strategy that has great benefits for tasks with high-dimensional observations. We use BiGAN to estimate the state novelty during training; then, we combine the state novelty as a weighted intrinsic reward with the extrinsic reward.
Under the resettable environment premise, we use an episodic memory-based method to handle the problem of vanishing intrinsic rewards. Our exploration strategy can be combined with any policy optimization algorithm.
Adventurer shows promising experimental results on continuous robotic manipulation tasks (e.g. Mujoco robotics) and high-dimensional image-based tasks (e.g. Atari games).

Efficient exploration is still an open problem as no method works better than all others in all environments \cite{Taiga2020On}.
Much \textbf{future research} is needed. First, while the concept of novelty is intuitive, it has yet to be rigorously defined as existing methods only propose proxies for novelty. Furthermore, while various conjectures exist, there still lacks a thorough analysis and deep understanding of when and why a particular method works well. In addition, how can existing methods be extended to work well in cases with extremely sparse rewards, e.g. a task with long-term delayed reward? Also, when resettability is not available, how can we automatically adjust the tradeoff between exploration and exploitation to address the vanishing intrinsic reward problem? 
Last, can we further combine various exploration techniques coherently, much like in ensemble learning, to produce a more robust performance under different scenarios? These are some of our future directions.















\appendix

\section{Resettable Premise}\label{sec:reset}
To solve the intrinsic reward vanishing problem, Adventurer uses episodic memory to store novel states and starts exploration from novel states at the beginning of each episode.
To realize it, we assume the environment is resettable, which means the agent can return to any given state easily without added exploration~\cite{ecoffet2021first}.

Under the resettable assumption, we build an episodic memory $M$ in each epoch to remember $K$ states with the highest novelty score $B(s)$, where $K$ is the memory size. The memory updates or adds a state when it is visited with a larger $B(s)$. Before the agent starts the exploration in each episode, it first chooses a novel initial state from the last epoch episodic memory and then returns to that state to start the exploration. The novel initial state enables the agent is more likely to explore more new states. 

We demonstrate how to incorporate the intrinsic reward and episodic memory into a full PPO agent in Algorithm~\ref{alg:adventurer_reset}. 
In each epoch, we update the BiGAN to fit the distribution of visited states and we also update the policy to exploit the interactions. For each episode, we first sample a novel initial state from the last epoch episodic memory if the environment is resettable. If not, we start with the initial state given by the environment. In each step, we calculate the state novelty score to do intrinsic-reward-based exploration.

In practice, we only need two episodic memories with $K$ length, since we only need the episodic memory from the last epoch where the agent can sample a novel initial state and the episodic memory in the current epoch which is used to save new novel states.
Moreover, when $l=0$, $M_{-1}$ does not exist, so we just sample a random initial state given by the environment for each episode. 

Moreover, compared to the memory in Go-explore~\cite{ecoffet2021first} which is based on number count, our episodic memory is more general, practical, and easy to implement. It can be integrated with novelty-driven methods easily. The visited number count method in Go-explore needs a lot of human engineering efforts which only works in some specific cases. Moreover, it squeezes the image observation which also imports unexpected count errors.

\begin{algorithm}[t]
	\caption{Adventurer: novelty-driven exploration} \label{alg:adventurer_reset}
	\begin{flushleft}
	Initialize policy network $\pi$ with parameters $\theta_{\pi}$;
	BiGAN network with parameters $\theta_{BiGAN}$;
	Episodic memory $M_{l}$, where $l$ is the epoch number;
	Novelty score reconstruction error scale $\alpha$;
	Exploration weight $\beta$;
	Total epoch number $L$;
	Total episodic number in each epoch $N$;
	Maximum step size for each episode $H$.
	\end{flushleft}
	
	\begin{algorithmic}[1]
	    \FOR {l = 0,...,L}
	    \FOR {n = 0,1,...,N}
	    \IF{resettable and l$\geq$ 1}
	    \STATE Sample a novel initial state $s_{0}$ from episodic memory $M_{l-1}$ and return to it.
	    \ELSE
	    \STATE Sample a random initial state $s_{0}$ from the environment.
	    \ENDIF
	    \FOR {h = 0,1,...,H}
	    \STATE Sample action $a_{t}\sim\pi(a_{t} \mid s_{t})$.
	    \STATE Sample $s_{t+1}, r^{e}_{t} \sim p(s_{t+1}, r^{e}_{t} \mid s_{t},a_{t})$.
	    \STATE Calculate novelty score $B(s_{t+1})$ with Eq.~\ref{eq:novel}.
	    \STATE Update episodic memory $M_{l}$ if needed.
	    \STATE Store $(s_{t},a_{t},r^{e}_{t},B(s_{t+1}),s_{t+1})$ into optimization buffer.
	    \ENDFOR
	    \ENDFOR
	     \STATE Update intrinsic reward normalization parameter $\mu(r^{e})$, $\mu(B(s))$, $\sigma(B(s))$.
	    \STATE Normalize $B(s_{t+1})$ into intrinsic reward $r^{i}_{t}$ with Eq.~5.
	    \STATE Calculate extrinsic reward advantage $A^{e}_{t}$ with Eq.~1.
	    \STATE Calculate intrinsic reward advantage $A^{i}_{t}$ with Eq.~1.
	    \STATE Calculate augmented advantage $A_{t} = A^{e}_{t} + \beta A^{i}_{t}$.
	    \STATE Update the policy network parameters $\theta_{\pi}$ with Eq.~2.
	    \STATE Update the BiGAN parameters $\theta_{BiGAN}$ on visited states with Eq.~3.
		\ENDFOR
	\end{algorithmic}
\end{algorithm}

\section{Extrinsic and intrinsic rewards}\label{sec:separate}

We learn two different advantage functions for extrinsic reward and intrinsic reward separately rather than directly modifying the reward in each step. We have some reasons why it is better to learn the rewards separately.
1) The sparseness of extrinsic reward and intrinsic reward is different. Usually, the extrinsic rewards are sparse in hard exploration environments and the intrinsic rewards are dense since we are calculating the novelty bonus in each step. We conjecture that learning mixed rewards is difﬁcult when the rewards are very different in nature.
2) The extrinsic reward function is stationary whereas the intrinsic reward is non-stationary since the intrinsic reward will change along with the updating of the novelty estimator.
3) By learning the rewards separately, we can use different discount factors for each type of reward which allows us to control the exploration bonus more precisely.

\section{Hyperparameters}\label{sec:hyper}
Most parameters work well with empirical value. We use the same hyperparameters for Adventurer and baselines except for $\alpha$, which is unique for Adventurer. It means our comparison is fair. Specially, we evaluate the impact of hyperparameter $\alpha$, where $\alpha$ equals to \{0.5, 0.7, 0.9, 1\} respectively. As stated in Sec.~\ref{sec:mr},  $L_{G}(s)$ performs better in detecting novel observations and $L_{D}(s)$ outperforms in reducing visited observations novelty score. Therefore, it is desirable to combine $L_{G}(s)$ and $L_{D}(s)$ together to  $B(s)$. We do a grid search on $\alpha$ with a range of \{0.5, 0.7, 0.9, 1\}.  The objective is to find the appropriate value to minimize  $D_{KL}(D_{1b} || D_{1a}) - D_{KL}(D_{2b} || D_{1a})$ in setting 1 and minimize $D_{KL}(D_{1b} || D_{1a}) + D_{KL}(D_{2b} || D_{2a})$ in setting 2. Finally, we choose $\alpha=0.9$ as the best parameters that overlap the distribution of visited states and distinguish the distribution of novel states.

The hyperparameter $\beta$ represents the exploration bonus weight that drives the agent to novel states. If the value is too small, the agent cannot get enough exploration, which leads to sampling inefficiency and being stuck in local optima. If the value is too large, the agent becomes biased toward the exploratory behavior which deviates from the extrinsic rewards. We first normalize the intrinsic rewards in E.q.~\ref{eq:normalization} and make sure that the extrinsic and intrinsic rewards are on the same scale by subtracting the mean of intrinsic rewards and adding the mean of extrinsic rewards. Then we do a grid search of  $\beta$ in the range of \{0.2, 0.3, 0.5\} and find that when $\beta=0.3$, we can get the best average performance. The hyperparameter $\beta$ affects the average performance but it is not sensitive that different $\beta$ shows relatively stable variant performance. Agent57~\cite{badia2020agent57} learns a family of policies with different hyperparameter combinations. It is possible to increase our performance if computation is enough.

\bibliographystyle{named}
\bibliography{ijcai22}

\end{document}